\documentclass{article}

\usepackage{PRIMEarxiv}
\usepackage{tikz}
\usepackage{amsmath}
\usepackage{algorithm}
\usepackage[noend]{algpseudocode}
\usepackage{float}
\usepackage{pgfplots}
\usepackage[utf8]{inputenc} 
\usepackage[T1]{fontenc}    
\usepackage{hyperref}       
\usepackage{url}            
\usepackage{booktabs}       
\usepackage{amsfonts}       
\usepackage{nicefrac}       
\usepackage{microtype}      
\usepackage{lipsum}
\usepackage{fancyhdr}       
\usepackage{graphicx}       
\graphicspath{{assets/}}     

\title{$Easter2.0$: Improving Convolutional models for Handwritten Text Recognition
}

\author{
  Kartik Chaudhary\\
  Data Scientist \\
  Bangalore, India\\
  \texttt{kartikgill@google.com} \\
   \And
  Raghav Bali \\
  Staff Data Scientist \\
  Berlin, Germany\\
  \texttt{raghav.bali@deliveryhero.com} \\
}

\begin{document}
\maketitle

\begin{abstract}
Convolutional Neural Networks (CNN) have shown promising results for the task of Handwritten Text Recognition (HTR) but they still fall behind Recurrent Neural Networks (RNNs)/Transformer based models in terms of performance. In this paper, we propose a CNN based architecture that bridges this gap. Our work, $Easter2.0$, is composed of multiple layers of 1D Convolution, Batch Normalization, ReLU, Dropout, Dense Residual connection, Squeeze-and-Excitation module and make use of Connectionist Temporal Classification (CTC) loss. In addition to the $Easter2.0$ architecture, we propose a simple and effective data augmentation technique 'Tiling and Corruption ($TACo$)' relevant for the task of HTR/OCR. Our work achieves state-of-the-art results on IAM handwriting database when trained using only publicly available training data. In our experiments, we also present the impact of $TACo$ augmentations and Squeeze-and-Excitation (SE) on text recognition accuracy. We further show that $Easter2.0$ is suitable for few-shot learning tasks and outperforms current best methods including Transformers when trained on limited amount of annotated data. Code and model is available at: \url{https://github.com/kartikgill/Easter2}.
\end{abstract}


\section{Introduction and Related Work}
Convolutional Neural Network (CNN) based models have been successfully applied in the field of Optical Character Recognition (OCR)\cite{Chaudhary2021EASTER,ingle2019scalable,bluche2017gated,ptucha2019intelligent,yousef2020accurate,chollet2017xception}. The key advantage of using a CNN based model is it's parameter efficiency and speed, however RNN/Transformer based architectures outperform them with a significant margin\cite{kang2020pay,diaz2021rethinking,li2021trocr,michael2019evaluating,pham2014dropout,voigtlaender2016handwriting}. Major difference between CNN and RNN/Transformer architectures is the access of context information. A typical CNN kernel with a limited kernel-size only attends to a small portion of input sequence/image, while the RNN cells have access to the context of entire input sequence (at least theoretically). Transformer based architectures use extensive attention layers to learn the global context. 

Understanding of such architectures and design choices led us to experiment with Squeeze-and-Excitation \cite{hu2018squeeze} ,for the task of text recognition, to introduce global context information in a CNN based architecture\cite{han2020contextnet} as well. Our experiments showcase how availability of global context improves model performance significantly.

\begin{figure}[H]
\centering
\begin{tikzpicture}
\begin{axis}[
scaled y ticks=real:1,
ytick scale label code/.code={},
ymax = 25,
symbolic x coords={20\%,40\%, 60\%,80\%, 100\%},
xtick=data,
height=7cm,
width=11cm,
grid=minor,
xlabel={IAM-Training Data Percentage (\%)},
ylabel={IAM-Test $CER$ (\%)},
legend style={
cells={anchor=east},
legend pos=north east,
}
]

\addplot coordinates {
(20\%, 19.32) (40\%, 14.14) (60\%,10.05) (80\%, 9.41) (100\%, 6.21) 
};

\addplot coordinates {
(20\%, 73.81) (40\%, 17.34) (60\%,10.14) (80\%, 10.11) (100\%, 7.62) 
};

\addplot coordinates {
(20\%, 20.61) (40\%, 16.15) (60\%,15.61) (80\%, 12.18) (100\%, 11.91) 
};

\addplot coordinates {
(100\%, 14.1) 
};

\addplot coordinates {
(100\%, 9.8) (60\%, 14.0)
};

\legend{$Easter2.0(Ours)$,$Transformer$\cite{kang2020pay},$Seq2Seq$\cite{kang2020pay},$GRCL$\cite{ingle2019scalable},$FCN/CTC$\cite{Chaudhary2021EASTER}}
\end{axis}
\end{tikzpicture}
     \caption{Few-shot training and performance comparison in terms of case-sensitive $CER$ with different portions of publicly available IAM-Training data.}
     \label{fig:comparison2}
\end{figure}
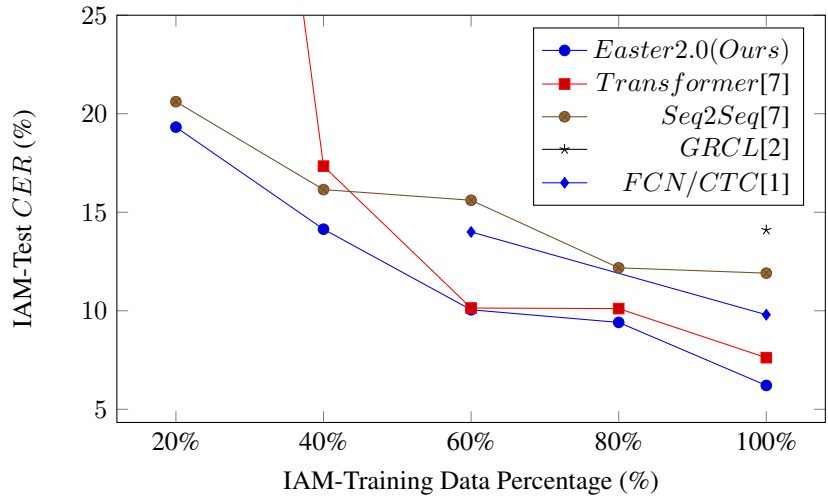

In the context of handwritten text recognition, very few works have used entirely convolutional architectures devoid of any recurrence \cite{Chaudhary2021EASTER,ptucha2019intelligent,yousef2020accurate}. Some recent works\cite{ingle2019scalable,coquenet2020recurrence} combine CNNs with a gating mechanism to compensate for the dependency on LSTM cells, known as Gated Convolutional Neural Networks (GCN). GCN architectures have been shown to outperform fully convolutional architectures yet they lag behind RNN/Transformer based architectures such as \cite{kang2020pay,diaz2021rethinking,li2021trocr}. 

Recurrent Neural Networks (RNN) work well for large variety of tasks including Handwritten Text Recognition (HTR). LSTM based models can handle long term context in sequences but at the cost of long learning times and potentially a huge number of parameters. Work by Voigtlaender et. al. \cite{voigtlaender2016handwriting} uses multi-dimensional LSTM to learn dependencies over both axis (horizontal and vertical), which makes them even slower.The most common architectures have a combination of CNN and RNN, where CNN is used for feature extraction from images and RNN is used for modeling sequential context \cite{wigington2017data,bluche2017gated}. Works by Michael and Labahn et. al. \cite{michael2019evaluating} and Kang and Toledo et. al.\cite{kang2018convolve} improve CNN+LSTM architectures with use of various attention mechanisms. While some works such as \cite{michael2019evaluating,bluche2016joint,chowdhury2018efficient,sueiras2018offline} have also experimented with sequence-to-sequence approaches. 

Recent progress in tasks associated with text recognition has achieved state-of-the-art (SOTA) results using Transformer-based architectures \cite{diaz2021rethinking,li2021trocr,kang2020pay}. Some of these works use a CNN-based backbone with self-attention as encoders to understand images\cite{li2021trocr}. Li and Lv et. al.\cite{li2021trocr} use pre-trained Computer Vision(CV) and Natural Language Processing (NLP) models to improve Transformer-based encoder-decoder architecture to achieve SOTA results on IAM handwriting dataset without an external language model. This approach leverages a large sized synthetic training set and multiple pre-trainings with a large number of training parameters.

We believe that the biggest challenge in developing an HTR system is not modeling but obtaining sufficient amounts of high quality training data. In this paper we address this problem by presenting a method that achieves encouraging results with very less amount of training samples. We also propose a CNN based architecture, $Easter2.0$,  that is data efficient, parameter-efficient, compute-efficient, latency-efficient and is easy to understand and deploy. Our model utilizes multiple layers of 1D Convolution, Batch Normalization, ReLU, Dropout, Dense Residual Connections, a Squeeze-and-Excitation (SE)\cite{hu2018squeeze} module and make use of  Connectionist Temporal Classification (CTC) loss\cite{graves2006connectionist}. The SE module improves access to global context for our proposed CNN architecture. $Easter2.0$ has advantages from both worlds, i.e., speed and parameter efficiency of CNN and access to global context similar to RNN/Transformers (using the SE module).  
The contributions of this paper are summarized as follows:
\begin{enumerate}
\item We propose $TACo$, a simple and novel data augmentation technique for Handwritten Text Recognition with experimental results to prove it's effectiveness.
\item We present $Easter2.0$, a novel CNN based architecture for the task of end-to-end handwritten text line recognition that utilizes only 1D Convolutions with dense residual connections and a squeeze-and-excitation module.
\item $Easter2.0$ is recurrence-free, parameter efficient, fast and simple convolutional architecture that achieves state-of-the-art results on IAM handwritten test set when trained using only publicly available data (IAM-Train set).
\end{enumerate}

\begin{figure}
  \centering
  \includegraphics[scale=0.4]{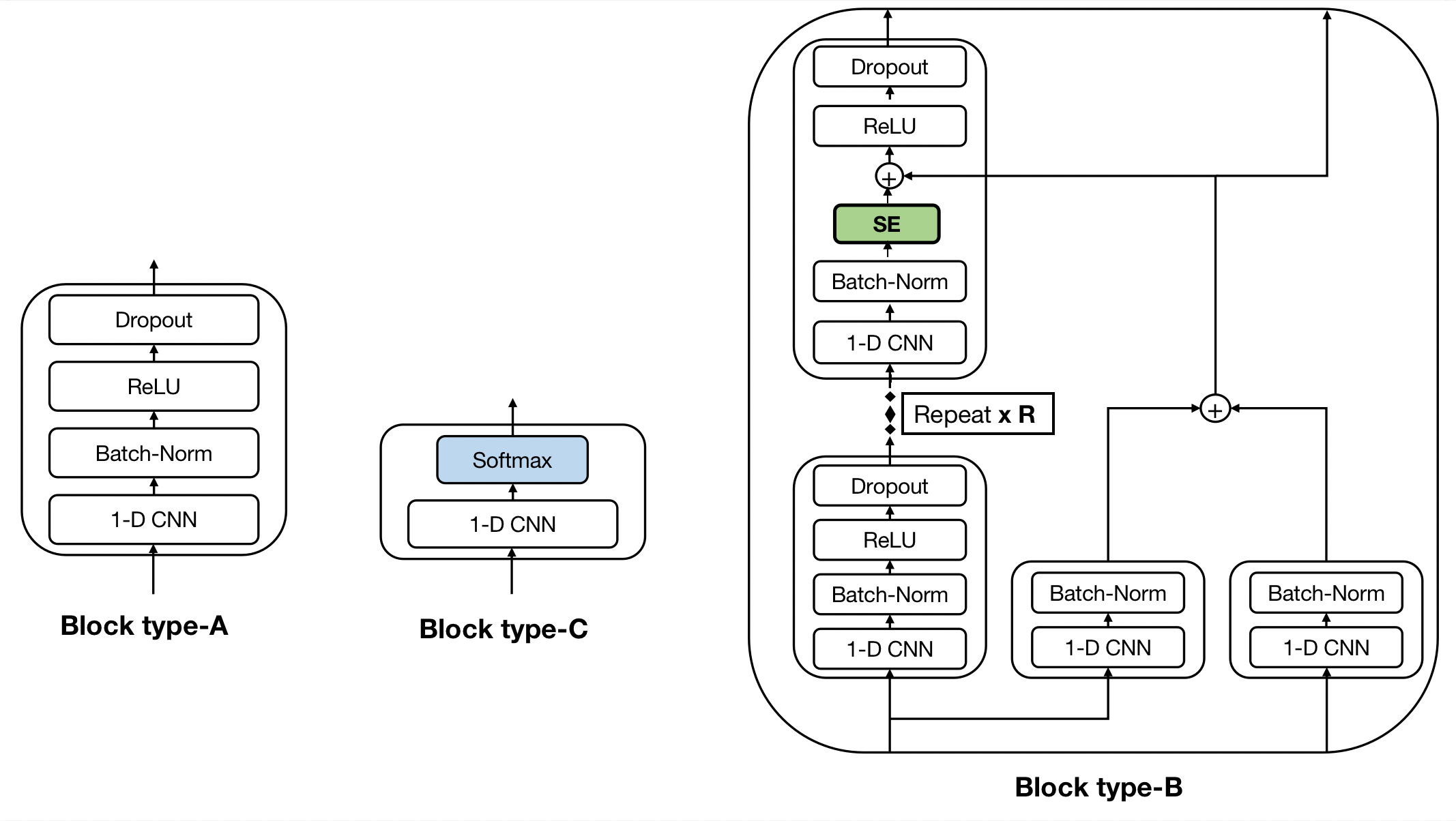}
  \caption{$Easter2.0$ Architecture: ($Left$) Block type-A has 1-DCNN, Batch Norm, ReLU and Dropout layers. ($Middle$) Block type-C has 1DCNN followed by a Softmax layer. ($Right$) Block type-B has repetitions of type-A blocks with Dense Residual Connections and SE module.}
  \label{fig:arch}
\end{figure}

\begin{figure}
  \centering
  \includegraphics[scale=0.25]{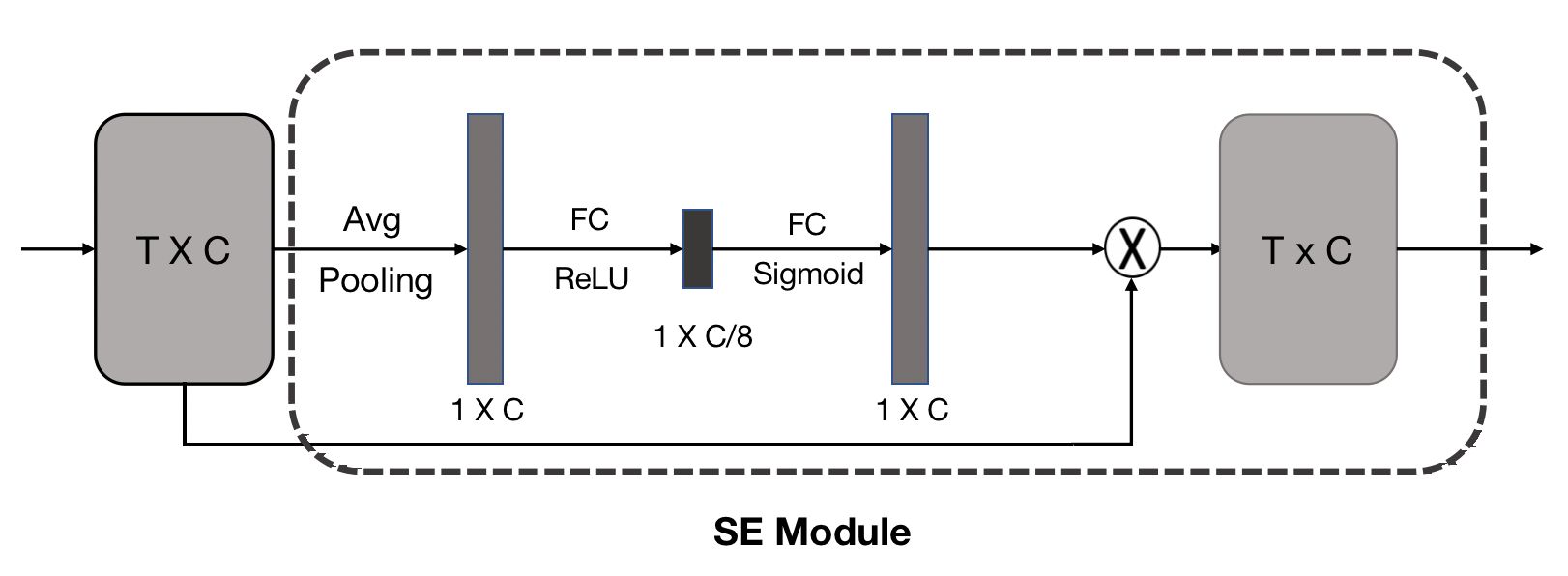}
  \caption{1D Squeeze-and-Excitation module applies a sequence of steps starting with average pooling on local features followed by two fully connected layers to get context vector. The context vector is multiplied element wise with local features.}
  \label{fig:se}
\end{figure}

\section{Tiling and Corruption Augmentation($TACo$)} \label{taco_section}
We introduce a simple and effective augmentation technique for the task of handwritten text line recognition. $TACo$ augmentations help the model to learn useful features and achieve good results even on very small training sets. 

\begin{figure}
  \centering
  \includegraphics[scale=0.25]{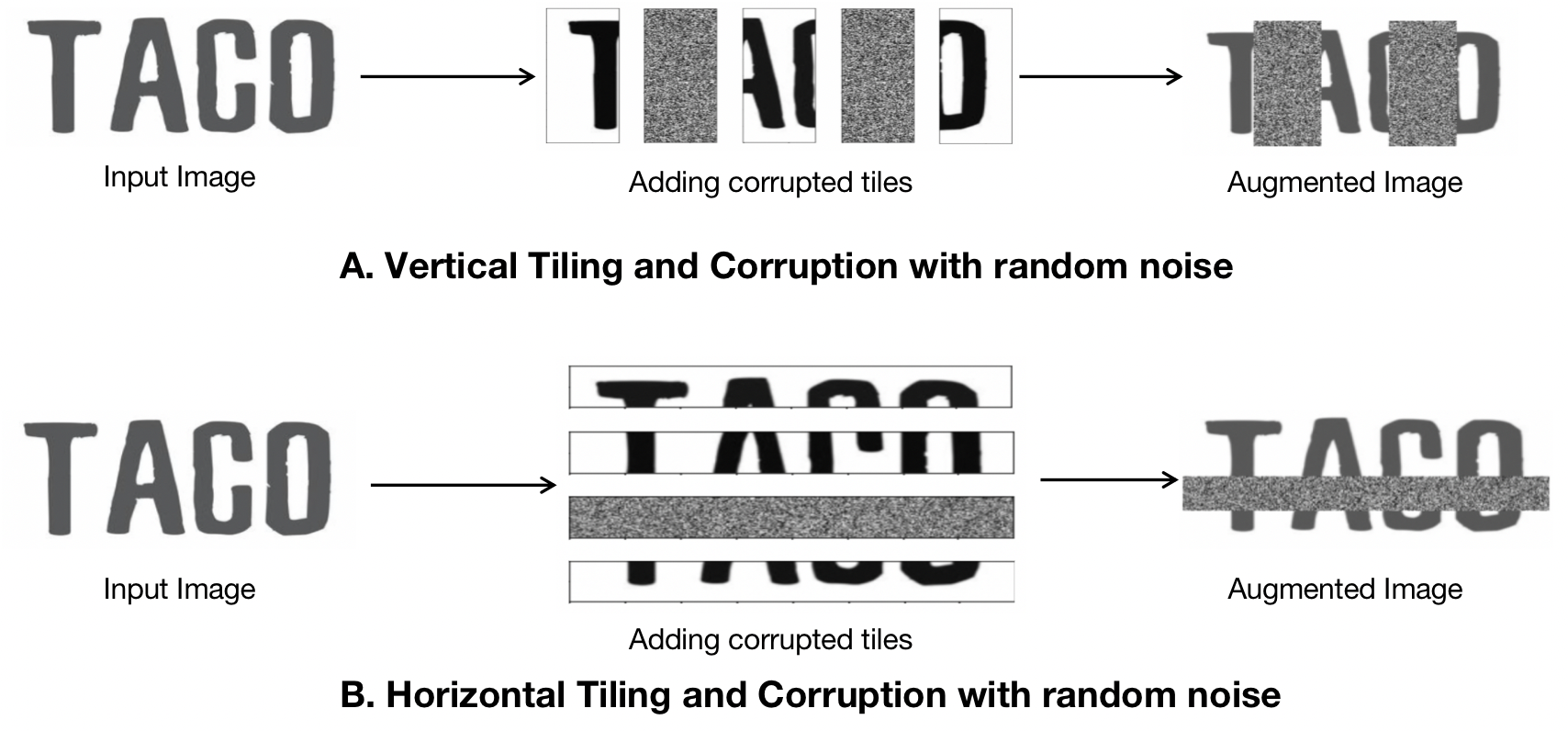}
  \caption{$TACo$ augmentation examples. ($Top$) Example of Vertical Tiling (across width) and Corruption with random noise. ($Bottom$) Shows Horizontal Tiling (across height) and Corruption with random noise.}
  \label{fig:TACo}
\end{figure}

Tiling refers to cutting the input image into multiple small tiles of equal size ($T_w$). After tiling, a fraction ($C_p$) of the tiles is replaced with corrupted ones as part of the corruption step ($C_p$ is a hyper-parameter used to set the corruption probability). Finally, tiles are stitched back together in the same order to get the augmented image(see figure \ref{fig:TACo}). Tile width $T_w$ is sampled from a uniform distribution from $\frac{H}{10}$ to maximum tile width parameter $T_{max}$ (where $H$ is height of input image). $TACo$ augmentations can be applied on an input image across height($H$), width($W$) or both. The $TACo$ augmentation algorithm is presented in algorithm-\ref{taco}:

\makeatletter
\def\BState{\State\hskip-\ALG@thistlm}
\makeatother

\begin{algorithm}
\caption{Tiling and Corruption ($TACo$)}\label{taco}
\begin{algorithmic}[1]
\Procedure{Apply$TACo$}{$img$}
\State $\textit{H, W} \gets \text{Shape of }\textit{img}$
\State $T_w \gets \text{sample from uniform distribution [$\frac{H}{10}$, $T_{max}$]}$
\State $\textit{tiles} \gets \text{cut and return a list of image tiles each with width = $T_w$}$

    \For {$i = 1 \to length(tiles)$}
        \State $\textit{$p$} \gets \text{random floating point number from range [0.0, 1.0)}$
        \Comment \emph{using pseudo-random number generator}
            \If {$p$ <= $C_p$}
                \Comment \emph{$C_p$:Corruption\ Probability}
                \State $\textit{$tiles[i]$} \gets \text{a corrupt tile of width ($T_w$) }$
                \Comment \emph{tile replacement}
            \EndIf
    \EndFor
    
\State $\textit{augmented\_img} \gets \text{join back tiles in the same order}$
\State \textbf{return} {$augmented\_img$}
\EndProcedure
\end{algorithmic}
\end{algorithm}

\begin{figure}
  \centering
  \includegraphics[scale=0.25]{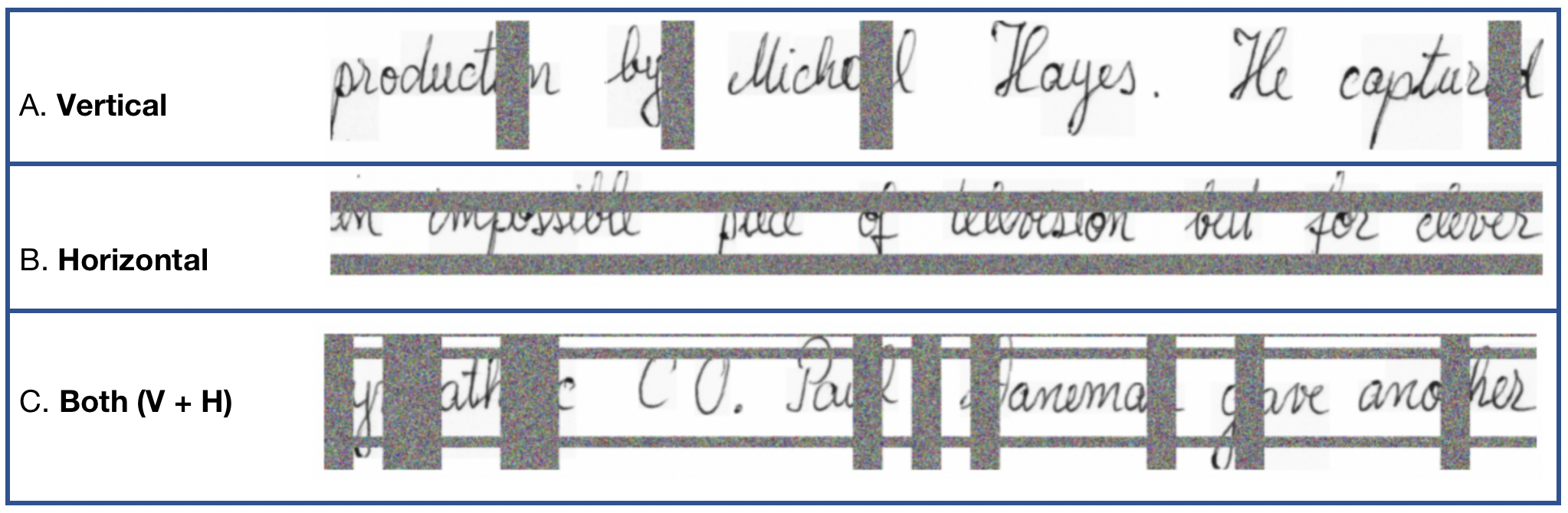}
  \caption{$TACo$ augmentation examples of IAM-Handwritten-lines. ($Top$) Example with only Vertical $TACo$. ($Center$) Represents Horizontal $TACo$. ($Bottom$) An example of hybrid (Vertical + Horizontal) $TACo$.}
  \label{fig:taco_iam}
\end{figure}

\section{Model Architecture}
\label{sec:headings}
The architecture of $Easter2.0$ is inspired by the design of $Easter$\cite{Chaudhary2021EASTER} which is a fully convolutional architecture with only 1D convolutions. However, there are some key differences in our proposed $Easter2.0$ architecture. $Easter2.0$ makes use of dense residual connections and a squeeze-and-excitation module to introduce global context into local features. Sections 3.1 to 3.3 describe components of our model in more detail and section 3.4 describes the final configuration of $Easter2.0$.

\subsection{Convolution Blocks}
$Easter2.0$ has overall 14 layers and each layer represents one of the 3 types of convolution blocks A, B or C as shown in figure \ref{fig:arch}). Block type-A is composed of a standard 1D convolutional layers followed by layers of batch normalization, ReLU and dropout. Block type-B has $R$ repetitions of Block type-A with a residual connection. The residual connection is first projected through a $1$ x $1$ convolution to balance the number of channels followed by a batch normalization layer. The output of this batch normalization layer is then added to the output of SE layer in the last convolution block, just before the layers of ReLU and Dropout (see figure  \ref{fig:arch}). The SE layer is only present in the last convolution block of Block type-B, just after the batch normalization layer(see figure \ref{fig:arch}). Block type-C is only used as the last layer of our model. Block type-C regulates size of output via $1$x$1$ convolution layers and is followed by a softmax layer to calculate the distribution of probabilities over the characters of given vocabulary.

\subsection{Dense Residual Connections}
$Easter2.0$ utilizes dense residual connections as shown in the Block type-B of figure \ref{fig:arch}. The dense residual connections\cite{li2019jasper} are obtained by adding the output of a convolution block to the inputs of all following blocks of type-B as shown in the figure \ref{fig:arch}. We found dense residual connections to be performing better than normal residual connections at the cost of small increase in training parameters (more details in the experiments section).

\subsection{Squeeze-and-Excitation}
We use a 1-D version of squeeze-and-excitation module to introduce global context as depicted in figure \ref{fig:se}. SE module squeezes the local features into a single global context vector of weights and broadcasts this context back to each local feature vector through element-wise multiplication of context weights with features. We present experimental study on how SE module improves the model performance. Checkout the pseudo-code for our SE module implementation \ref{se} which is very similar to the works of Hu and Shen et. al.\cite{hu2018squeeze} and Han and Zhang et. al. \cite{han2020contextnet}.

\begin{algorithm}
\caption{Squeeze-and-Excitation-1D($SE$)}\label{se}
\begin{algorithmic}[1]
\Procedure{Add\_Global\_Context}{$data$,$filters$}
\State $\textit{x} \gets \text{GlobalAveragePooling1D on }\textit{\ data}$
\State $\textit{x} \gets \text{FullyConnected (units=$\frac{filters}{8}$) on }\textit{\ x}$
\Comment \emph{Bottleneck}
\State $\textit{x} \gets \text{ReLU on }\textit{\ x}$
\State $\textit{x} \gets \text{FullyConnected (units=$filters$) on}\textit{\ x}$
\State $\textit{weights} \gets \text{Sigmoid on }\textit{\ x}$
\State $\textit{final\_data} \gets \text{Element-wise Multiplication ($weights$, $data$)}$
\Comment \emph{context broadcasted}
\State \textbf{return} {$final\_data$}
\EndProcedure
\end{algorithmic}
\end{algorithm}

\subsection{Configuration}
$Easter2.0$ has a total $14$ blocks of convolution layers. Table \ref{tab:config_table} summarizes the architectural details. First two layers are type-A blocks (B1 and B2) with kernel width of $3$ and a stride of $2$. Next $3$ blocks are type-B blocks (B3 to B5) with kernel widths of $5$, $7$ and $9$ respectively with dense residual connections. Blocks B6 and B7 are again type-A blocks with kernel widths of $11$ and $1$ respectively. Block B6 has a dilation rate of $2$. Block B8 is a type-C block. All blocks except B8, have dropout layers. 

\begin{table}
 \caption{$Easter2.0$ configuration. The kernel size is for the window size across width and convolutions are across height.}
  \centering
  \begin{tabular}{lllllll}
    \toprule
      Block & Block & Conv & \#Output & Kernel & Other & Dropout \\
      ID & type  & layers & Channels & Size &   & \\
    \midrule
    B1,B2 & A & 1 & 128 & 3 & stride=2 & 0.2\\
    B3 & B & 3 & 256 & 5 & dense residual & 0.2\\
    B4 & B & 3 & 256 & 7 & dense residual & 0.2\\
    B5 & B & 3 & 256 & 9 & dense residual & 0.3\\
    B6 & A & 1 & 512 & 11 & dilation=2 & 0.4\\
    B7 & A & 1 & 512 & 1 & - & 0.4\\
    B8 & C & 1 & \#Vocab & 1 & - & -\\
    \bottomrule
  \end{tabular}
  \label{tab:config_table}
\end{table}

\section{Experiments}
\label{sec:headings2}
\subsection{Dataset}
The biggest challenge for OCR/HTR tasks is obtaining good quality labelled data
IAM handwriting database\cite{marti2002iam} is a popular benchmark for comparing HTR models. Recent studies have utilised synthetic/internal datasets in addition to IAM-Training set to obtain SOTA results on IAM-Test set. Curation and usage of such synthetic datasets presents difficulties in reproducibility. We only make use of publicly available datasets for our experiments.

\subsubsection{IAM} \label{iam}The IAM handwritten dataset is composed of 1,539 scanned
form pages from 657 writers. It corresponds to handwritten English text images also available as grayscaled images. There are 79 characters in the alphabet. We only focus on the line level dataset. \textbf{IAM-A} has about $6,482$, $976$ and $2,915$ lines for training, validation and test datasets respectively. \textbf{IAM-B} has about $6,161$, $940$ and $1,861$ lines for training, validation and test datasets respectively. 
\subsubsection{Long-Lines} \label{long-lines}
Long line samples are obtained by randomly choosing two training images and stacking them horizontally (with a small white background image in between) to form a long line and labels are concatenated with a space. This introduces examples with long lengths, multiple handwritings, strokes, etc. 

\subsection{Training Details}
The experiments are carried out with Tensorflow\cite{abadi2016tensorflow} toolkit with a weighted Connectionist Temporal Classification (w-CTC)(similar to \cite{Chaudhary2021EASTER}) as loss function. We have used Adam optimizer with an initial learning rate of $10^{-3}$ and a batch size of $32$ with early stopping criteria. 
\subsection{Evaluation}
We use case-sensitive Character Error Rate ($CER$) as evaluation metric to compare results in our experiments. We also provide model parameters from our experiments. The $CER$ is computed as the Levenshtein distance which is the
sum of the character substitutions ($Sc$), insertions ($Ic$) and
deletions ($Dc$) that are needed to transform one string into
the other, divided by the total number of characters in the
groundtruth ($Nc$). Formally,

\begin{equation}
    CER(\%) = {\frac{(Sc + Ic + Dc)}{Nc}}\times{100}
\end{equation}

\subsection{Results}
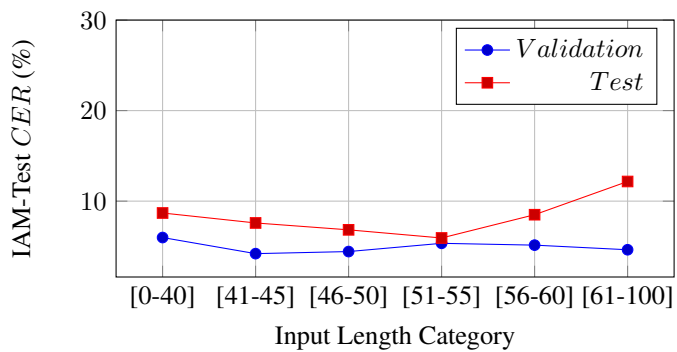
\begin{figure}[H]
\centering
\begin{tikzpicture}
\begin{axis}[
scaled y ticks=real:1,
ytick scale label code/.code={},
ymax = 30,
symbolic x coords={[0-40],[41-45],[46-50],[51-55],[56-60],[61-100]},
xtick=data,
height=5cm,
width=9cm,
grid=major,
xlabel={Input Length Category},
ylabel={IAM-Test $CER$ (\%)},
legend style={
cells={anchor=east},
legend pos=north east,
}
]

\addplot coordinates {
([0-40], 5.98) ([41-45], 4.20) ([46-50], 4.43) ([51-55], 5.34) ([56-60], 5.14) ([61-100], 4.63) 
};

\addplot coordinates {
([0-40], 8.69) ([41-45], 7.59) ([46-50],6.83) ([51-55], 5.93) ([56-60], 8.5) ([61-100], 12.17) 
};

\legend{$Validation$,$Test$}
\end{axis}
\end{tikzpicture}
     \caption{Effect of Input Length on model performance (in terms of $CER$).}
     \label{fig:comparison1}
\end{figure}

In this section, we present results for different experiments based on different configurations discussed so far.
\begin{enumerate}
    \item \textbf{Effect of Residual Connections: }
    Residual connections improve training when networks are deep. Table \ref{tab:res_se_table} shows the effect of residual connections on performance. We found out that dense residual connections outperform normal residual connections by a good margin. 
    \item \textbf{Effect of Normalization: }
    In our study, we evaluate performance of our model with 2 normalization techniques: batch normalization\cite{ioffe2015batch} and layer normalization\cite{ba2016layer} and also without normalization. Experiment results are shown in Table \ref{tab:norm_table}. We found that batch normalization outperforms layer normalization. Both normalization techniques help model train faster and achieve better results than the case without normalization.
    \item \textbf{Effect of Squeeze-and-Excitation: }
    The squeeze-and-excitation module introduces the global context into convolutions. We found that adding SE to our network improves the model performance with a small increase in parameters. Table \ref{tab:res_se_table} shows the accuracy improvements when SE module is applied to $Easter2.0$ with dense as well as normal residual connections.
    \item \textbf{Effect of $TACo$ Augmentations: }
    $TACo$ being a simple augmentation technique improves the model performance considerably. Table \ref{tab:taco_table} shows that $TACo$ improves $CER$ irrespective of the type of corruption applied to certain parts of the image. We finally went ahead with random-noise as a type of corruption in rest of our experiments.
    \item \textbf{Effect of Tile Width: }
    We experimented with different values of maximum tile widths ($T_{max}$) while applying $TACo$ augmentations (check section \ref{taco_section}). In our experiments, we found out that $TACo$ improves $Easter2.0$ irrespective of tile widths (see Table \ref{tab:taco_size_table}). Model achieves best results when $T_{max}$=$H$ ($H$ is height of input image).
    \item \textbf{Effect of Input Length: }
    Our experiments suggest that $Easter2.0$ works well irrespective of input length and thus a good option for recognizing longer sequences. Figure \ref{fig:comparison1} shows the $CER$ analysis wrt. the input length on IAM-offline test set. 
    \item \textbf{Few-shot Training: }
    $Easter2.0$ is ideal for few-shot training and produces state-of-the-art (SOTA) results with very small amount of annotated data. Table \ref{tab:fs_table} shows some experiments that we conducted with limited number of training samples. We found out that $Easter2.0$ outperforms SOTA works including Transformers\cite{kang2020pay} when training data is small (see tables \ref{tab:fs_table} and \ref{tab:sota_table}). As obtaining large amount of good quality annotated data is a challenge in many recognition tasks, Our solution can be a choice in such scenarios. External language model can improve our model further however we have only presented greedy decoding results in this paper.
    \item \textbf{Comparison with the State-of-the-Art: }
    Table \ref{tab:sota_table} compares current state-of-the-art results on IAM-Test set. $Easter2.0$ beats previous SOTA results with a good margin when only publicly available data is used for training. Many state-of-the-art (SOTA) solutions use pre-trainings and large internal/synthetic datasets to boost model accuracy. As everyone has their own way to collect synthetic/internal data, it makes the model comparison unfair. To avoid this bias, we train $Easter2.0$ only with publicly available IAM-offline training set (6,482 lines). We present SOTA results on IAM-test set having 2,915 lines (see table \ref{tab:sota_table}). However, when synthetic/Internal data is allowed Diaz et al.,\cite{diaz2021rethinking} and TrOCR\cite{li2021trocr} achieve SOTA results on IAM-Test set with CER 2.75 and 2.89 respectively.
\end{enumerate}

\begin{table}
 \caption{Effects of Residual Connections and SE module evaluated in terms of case-sensitive Character Error Rate (CER) on IAM-Test set. This experiment uses IAM-B partition (see section \ref{iam})}
  \centering
  \begin{tabular}{llll}
    \toprule
      Residual Type & SE layer & CER(\%) & \#param \\
    \midrule
        - & - & 10.47 & 5.8M \\
        Normal & - & 10.24 & 5.9M \\
        Dense  & - & \textbf{9.18} & 6.1M \\
        Normal & \checkmark & 8.95 & 6.0M \\
        Dense  & \checkmark & \textbf{8.90} & 6.1M \\
    \bottomrule
  \end{tabular}
  \label{tab:res_se_table}
\end{table}


\begin{table}
 \caption{Effect of Normalization choices evaluated in terms of case-sensitive Character Error Rate (CER) on IAM-Test set. This experiment uses IAM-A partition (see section \ref{iam})}
  \centering
  \begin{tabular}{ll}
    \toprule
      Normalization Type & CER(\%)\\
    \midrule
        - & 11.42 \\
        Layer Norm & 9.05\\
        Batch Norm & \textbf{8.73}\\

    \bottomrule
  \end{tabular}
  \label{tab:norm_table}
\end{table}



\begin{table}
 \caption{Effect of $TACo$ augmentation-variations evaluated in terms of case-sensitive Character Error Rate (CER) on IAM-Test set. This experiment uses IAM-B partition (see section \ref{iam})}
 
  \centering
  \begin{tabular}{llll}
    \toprule
      \textbf{Corruption Type} & \textbf{Vertical} & \textbf{Horizontal} & \textbf{CER(\%)} \\
    \midrule
        - & - & - & 8.90 \\
    \midrule
        Black & \checkmark & - & 7.80 \\
              & \checkmark & \checkmark & 8.21 \\
    \midrule
        White & \checkmark  & - & 8.29 \\
              & \checkmark  & \checkmark & 8.09 \\
    \midrule
        Mean & \checkmark  & - & 7.78 \\
             & \checkmark  & \checkmark & 7.77 \\
    \midrule
        Random  & \checkmark  & - & \textbf{7.76} \\
                & \checkmark  & \checkmark & \textbf{7.72} \\
    \midrule
        Miscellaneous  & \checkmark  & - & 8.32 \\
    \bottomrule
  \end{tabular}
  \label{tab:taco_table}
\end{table}


\begin{table}
 \caption{Effect of $TACo$-Tile width evaluated in terms of case-sensitive Character Error Rate (CER) on IAM-Test set. This experiment uses IAM-A partition (see section \ref{iam})}
  \centering
  \begin{tabular}{llll}
    \toprule
      \textbf{Corruption} & \textbf{Tile Width} & \textbf{CER(\%)}\\
    \midrule
        - & - & 10.71 \\
        random, vertical & $H/2$ & 8.92 \\
        random, vertical  & $H$ & \textbf{8.73} \\
        random, vertical & $2\times H$ & 9.33 \\
        random, vertical  & $4\times H$ & 9.37 \\
    \bottomrule
  \end{tabular}
  \label{tab:taco_size_table}
\end{table}


\begin{table}
 \caption{Few-shot training and results comparison with SOTA solutions evaluated in terms of case-sensitive Character Error Rate (CER) on IAM-Test set. This experiment uses IAM-A partition (see section \ref{iam})}
  \centering
  \begin{tabular}{l*{6}{c}ll}
    \toprule
    Method & \multicolumn{6}{c}{$CER$ (\%)}& Time (s) & \#param (M)\\
    \cmidrule(lr){2-7}
    & \multicolumn{1}{c}{10\%} & \multicolumn{1}{c}{20\%} & 
      \multicolumn{1}{c}{40\%} & \multicolumn{1}{c}{60\%} &
      \multicolumn{1}{c}{80\%} & \multicolumn{1}{c}{100\%} & & \\
      
    \midrule
    Seq2Seq\cite{kang2020pay} & - & 20.61 & 16.15 & 15.61 & 12.18 & 11.91 & 338.7 & 37\\
    Transformer w/CNN\cite{kang2020pay} & - & 73.81 & 17.34 & 10.14 & 10.11 & \textbf{7.62} & 202.5 & 100\\
    \midrule
    $Easter2.0$ (Ours) & \textbf{38.90} & \textbf{19.32} & \textbf{14.14} & \textbf{10.05} & \textbf{9.41} & 8.73 & \textbf{51} & \textbf{6.1}\\
    $Easter2.0$ + Long-Lines(\ref{long-lines}) & - & - & - & - & - & \textbf{6.21} & \textbf{51} & \textbf{6.1}\\
    \bottomrule
  \end{tabular}
  \label{tab:fs_table}
\end{table}

\begin{table}[ht!]
 \caption{Comparison with previous solutions using case-sensitive CER on IAM-Test set (line-level test set with greedy decoding). These results use IAM-A partition (see section \ref{iam})}
  \centering
  \begin{tabular}{llllll}
    \toprule
      \textbf{Training Data} & \textbf{Model} & \textbf{Architecture} & \textbf{External LM} & \textbf{CER(\%)} & \textbf{\#param(M)} \\
    \midrule
        \textbf{IAM} (6k) & Ingle et al., 2019\cite{ingle2019scalable} & GRCL & No & 14.1 & 10.6 \\
        & Chaudhary et al.,2021\cite{Chaudhary2021EASTER} & FCN/CTC & No & 9.8 & 28 \\
        & Kang et al., 2020\cite{kang2020pay} & Transformer w/ CNN & No & 7.62 & 100 \\
        & Wang et al.\cite{wang2020decoupled} & DAN, FCN/GRU & No & 6.4 & - \\
        & \textbf{$Easter2.0$ (Ours)} & \textbf{1DCNN-SE/ CTC} & \textbf{No} & \textbf{6.21} & \textbf{6.1} \\
    \hline
    \midrule
    \textbf{Synthetic} + IAM & Kang et al., 2020 \cite{kang2020pay}& Transformer w/ CNN & No & 4.67 & 100 \\
    &Bluche and Messina, 2017\cite{bluche2017gated} & GCRNN / CTC & Yes & 3.2 & - \\
     & TrOCR(LARGE) 2021\cite{li2021trocr} & Transformer & No & \textbf{2.89} & 558 \\
     \hline
    \midrule
        \textbf{Internal} + IAM & Michael et al., 2019\cite{michael2019evaluating} & LSTM/LSTM w/Attn & No & 4.87 & - \\
        & Ingle et al., 2019\cite{ingle2019scalable}& GRCL & No & 4.0 & 10.6 \\
        & Diaz et al., 2021\cite{diaz2021rethinking} & Transformer w/ CNN & No & 2.96 & - \\
        & Diaz et al., 2021\cite{diaz2021rethinking} & S-Attn / CTC & Yes & \textbf{2.75} & - \\
    \bottomrule
  \end{tabular}
  \label{tab:sota_table}
\end{table}


\section{Conclusion}
In this paper, we proposed a convolutional architecture for the task of handwritten text recognition that utilizes only 1D convolutions, dense residual connections and a SE module. We also proposed a simple and effective data augmentation technique-$TACo$ useful for OCR/HTR tasks. We have presented experimental study on components of $Easter2.0$ architecture including dense residual connections, normalization choices, SE module, TACo variations and few-shot training. Our work achieves SOTA results on IAM-Test set when training data is limited, also $Easter2.0$ has very small number of trainable parameters compared to other solutions. The proposed architecture can be used in search of smaller, faster and efficient OCR/HTR solutions when available annotated data is limited.
\label{sec:headings_3}

\bibliographystyle{unsrt}  
\bibliography{references}  

\begin{thebibliography}{10}

\bibitem{Chaudhary2021EASTER}
Kartik Chaudhary and Raghav Bali.
\newblock Easter: Simplifying text recognition using only 1d convolutions.
\newblock {\em Proceedings of the Canadian Conference on Artificial
  Intelligence}, 6 2021.
\newblock https://caiac.pubpub.org/pub/fm5sy88o.

\bibitem{ingle2019scalable}
R~Reeve Ingle, Yasuhisa Fujii, Thomas Deselaers, Jonathan Baccash, and Ashok~C
  Popat.
\newblock A scalable handwritten text recognition system.
\newblock In {\em 2019 International Conference on Document Analysis and
  Recognition (ICDAR)}, pages 17--24. IEEE, 2019.

\bibitem{bluche2017gated}
Th{\'e}odore Bluche and Ronaldo Messina.
\newblock Gated convolutional recurrent neural networks for multilingual
  handwriting recognition.
\newblock In {\em 2017 14th IAPR international conference on document analysis
  and recognition (ICDAR)}, volume~1, pages 646--651. IEEE, 2017.

\bibitem{ptucha2019intelligent}
Raymond Ptucha, Felipe~Petroski Such, Suhas Pillai, Frank Brockler, Vatsala
  Singh, and Paul Hutkowski.
\newblock Intelligent character recognition using fully convolutional neural
  networks.
\newblock {\em Pattern recognition}, 88:604--613, 2019.

\bibitem{yousef2020accurate}
Mohamed Yousef, Khaled~F Hussain, and Usama~S Mohammed.
\newblock Accurate, data-efficient, unconstrained text recognition with
  convolutional neural networks.
\newblock {\em Pattern Recognition}, 108:107482, 2020.

\bibitem{chollet2017xception}
Fran{\c{c}}ois Chollet.
\newblock Xception: Deep learning with depthwise separable convolutions.
\newblock In {\em Proceedings of the IEEE conference on computer vision and
  pattern recognition}, pages 1251--1258, 2017.

\bibitem{kang2020pay}
Lei Kang, Pau Riba, Mar{\c{c}}al Rusi{\~n}ol, Alicia Forn{\'e}s, and Mauricio
  Villegas.
\newblock Pay attention to what you read: Non-recurrent handwritten text-line
  recognition.
\newblock {\em arXiv preprint arXiv:2005.13044}, 2020.

\bibitem{diaz2021rethinking}
Daniel~Hernandez Diaz, Siyang Qin, Reeve Ingle, Yasuhisa Fujii, and Alessandro
  Bissacco.
\newblock Rethinking text line recognition models.
\newblock {\em arXiv preprint arXiv:2104.07787}, 2021.

\bibitem{li2021trocr}
Minghao Li, Tengchao Lv, Lei Cui, Yijuan Lu, Dinei Florencio, Cha Zhang,
  Zhoujun Li, and Furu Wei.
\newblock Trocr: Transformer-based optical character recognition with
  pre-trained models.
\newblock {\em arXiv preprint arXiv:2109.10282}, 2021.

\bibitem{michael2019evaluating}
Johannes Michael, Roger Labahn, Tobias Gr{\"u}ning, and Jochen Z{\"o}llner.
\newblock Evaluating sequence-to-sequence models for handwritten text
  recognition.
\newblock In {\em 2019 International Conference on Document Analysis and
  Recognition (ICDAR)}, pages 1286--1293. IEEE, 2019.

\bibitem{pham2014dropout}
Vu~Pham, Th{\'e}odore Bluche, Christopher Kermorvant, and J{\'e}r{\^o}me
  Louradour.
\newblock Dropout improves recurrent neural networks for handwriting
  recognition.
\newblock In {\em 2014 14th international conference on frontiers in
  handwriting recognition}, pages 285--290. IEEE, 2014.

\bibitem{voigtlaender2016handwriting}
Paul Voigtlaender, Patrick Doetsch, and Hermann Ney.
\newblock Handwriting recognition with large multidimensional long short-term
  memory recurrent neural networks.
\newblock In {\em 2016 15th International Conference on Frontiers in
  Handwriting Recognition (ICFHR)}, pages 228--233. IEEE, 2016.

\bibitem{hu2018squeeze}
Jie Hu, Li~Shen, and Gang Sun.
\newblock Squeeze-and-excitation networks.
\newblock In {\em Proceedings of the IEEE conference on computer vision and
  pattern recognition}, pages 7132--7141, 2018.

\bibitem{han2020contextnet}
Wei Han, Zhengdong Zhang, Yu~Zhang, Jiahui Yu, Chung-Cheng Chiu, James Qin,
  Anmol Gulati, Ruoming Pang, and Yonghui Wu.
\newblock Contextnet: Improving convolutional neural networks for automatic
  speech recognition with global context.
\newblock {\em arXiv preprint arXiv:2005.03191}, 2020.

\bibitem{coquenet2020recurrence}
Denis Coquenet, Cl{\'e}ment Chatelain, and Thierry Paquet.
\newblock Recurrence-free unconstrained handwritten text recognition using
  gated fully convolutional network.
\newblock In {\em 2020 17th International Conference on Frontiers in
  Handwriting Recognition (ICFHR)}, pages 19--24. IEEE, 2020.

\bibitem{wigington2017data}
Curtis Wigington, Seth Stewart, Brian Davis, Bill Barrett, Brian Price, and
  Scott Cohen.
\newblock Data augmentation for recognition of handwritten words and lines
  using a cnn-lstm network.
\newblock In {\em 2017 14th IAPR International Conference on Document Analysis
  and Recognition (ICDAR)}, volume~1, pages 639--645. IEEE, 2017.

\bibitem{kang2018convolve}
Lei Kang, J~Ignacio Toledo, Pau Riba, Mauricio Villegas, Alicia Forn{\'e}s, and
  Mar{\c{c}}al Rusinol.
\newblock Convolve, attend and spell: An attention-based sequence-to-sequence
  model for handwritten word recognition.
\newblock In {\em German Conference on Pattern Recognition}, pages 459--472.
  Springer, 2018.

\bibitem{bluche2016joint}
Th{\'e}odore Bluche.
\newblock Joint line segmentation and transcription for end-to-end handwritten
  paragraph recognition.
\newblock {\em Advances in Neural Information Processing Systems}, 29:838--846,
  2016.

\bibitem{chowdhury2018efficient}
Arindam Chowdhury and Lovekesh Vig.
\newblock An efficient end-to-end neural model for handwritten text
  recognition.
\newblock {\em arXiv preprint arXiv:1807.07965}, 2018.

\bibitem{sueiras2018offline}
Jorge Sueiras, Victoria Ruiz, Angel Sanchez, and Jose~F Velez.
\newblock Offline continuous handwriting recognition using sequence to sequence
  neural networks.
\newblock {\em Neurocomputing}, 289:119--128, 2018.

\bibitem{graves2006connectionist}
Alex Graves, Santiago Fern{\'a}ndez, Faustino Gomez, and J{\"u}rgen
  Schmidhuber.
\newblock Connectionist temporal classification: labelling unsegmented sequence
  data with recurrent neural networks.
\newblock In {\em Proceedings of the 23rd international conference on Machine
  learning}, pages 369--376, 2006.

\bibitem{li2019jasper}
Jason Li, Vitaly Lavrukhin, Boris Ginsburg, Ryan Leary, Oleksii Kuchaiev,
  Jonathan~M Cohen, Huyen Nguyen, and Ravi~Teja Gadde.
\newblock Jasper: An end-to-end convolutional neural acoustic model.
\newblock {\em arXiv preprint arXiv:1904.03288}, 2019.

\bibitem{marti2002iam}
U-V Marti and Horst Bunke.
\newblock The iam-database: an english sentence database for offline
  handwriting recognition.
\newblock {\em International Journal on Document Analysis and Recognition},
  5(1):39--46, 2002.

\bibitem{abadi2016tensorflow}
Mart{\'\i}n Abadi, Paul Barham, Jianmin Chen, Zhifeng Chen, Andy Davis, Jeffrey
  Dean, Matthieu Devin, Sanjay Ghemawat, Geoffrey Irving, Michael Isard, et~al.
\newblock Tensorflow: A system for large-scale machine learning.
\newblock In {\em 12th $\{$USENIX$\}$ symposium on operating systems design and
  implementation ($\{$OSDI$\}$ 16)}, pages 265--283, 2016.

\bibitem{ioffe2015batch}
Sergey Ioffe and Christian Szegedy.
\newblock Batch normalization: Accelerating deep network training by reducing
  internal covariate shift.
\newblock In {\em International conference on machine learning}, pages
  448--456. PMLR, 2015.

\bibitem{ba2016layer}
Jimmy~Lei Ba, Jamie~Ryan Kiros, and Geoffrey~E Hinton.
\newblock Layer normalization.
\newblock {\em arXiv preprint arXiv:1607.06450}, 2016.

\bibitem{wang2020decoupled}
Tianwei Wang, Yuanzhi Zhu, Lianwen Jin, Canjie Luo, Xiaoxue Chen, Yaqiang Wu,
  Qianying Wang, and Mingxiang Cai.
\newblock Decoupled attention network for text recognition.
\newblock In {\em Proceedings of the AAAI Conference on Artificial
  Intelligence}, volume~34, pages 12216--12224, 2020.

\end{thebibliography}

\newpage
\appendix
\onecolumn

\section{APPENDIX- Easy Examples}
\begin{figure}[h]
  \centering
  \includegraphics[scale=0.45]{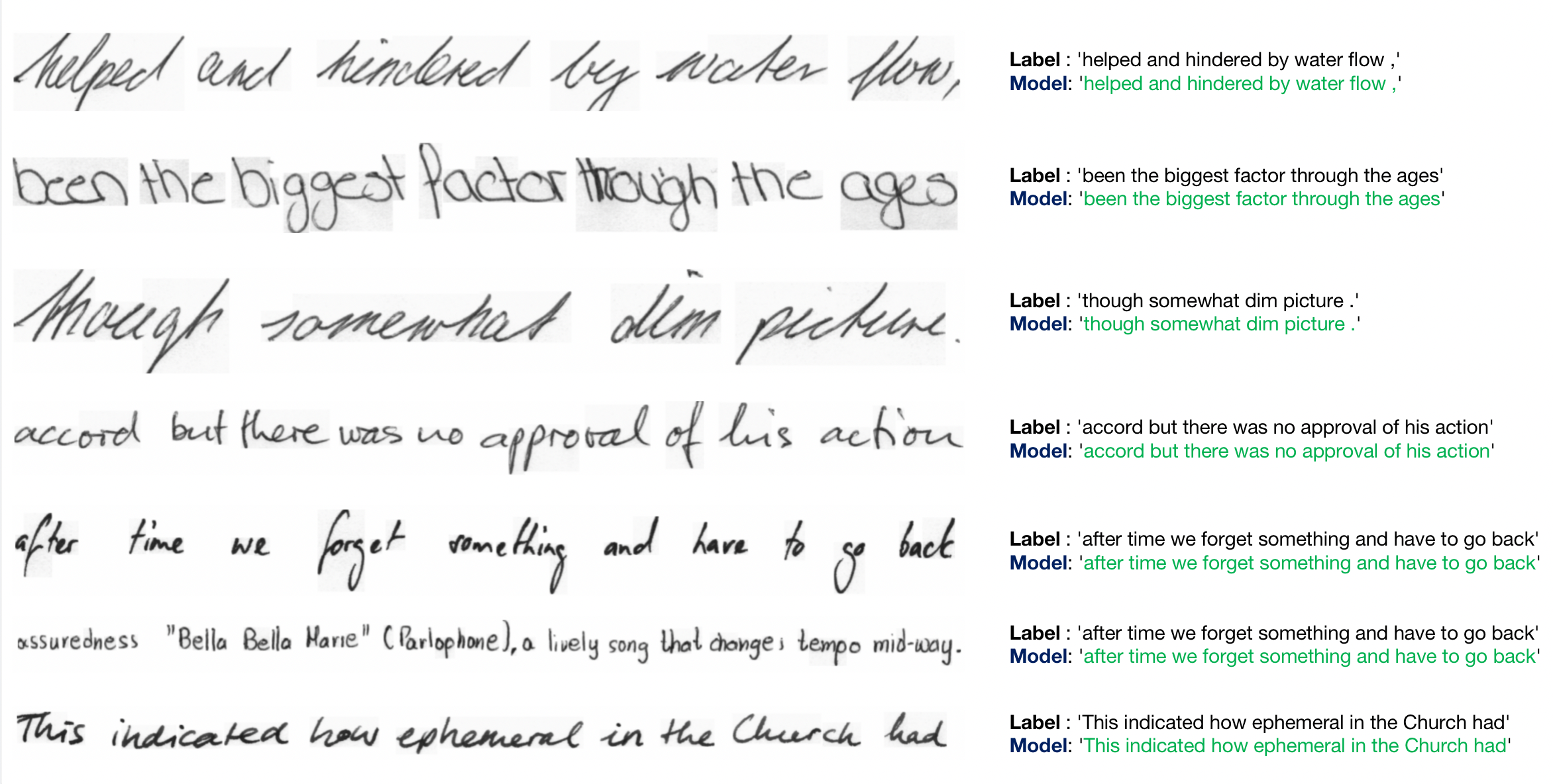}
  \caption{Examples where model doesn't make any mistake}
  \label{fig:easy}
\end{figure}

Figure \ref{fig:easy} shows some clean examples (From IAM-Offline Test partition) that were easy for model. In these examples, Easter2.0 doesn't make any mistake and recognises full sentence correctly.

\newpage

\section{APPENDIX- Difficult Examples}
\begin{figure*}[t]
  \centering
  \includegraphics[scale=0.45]{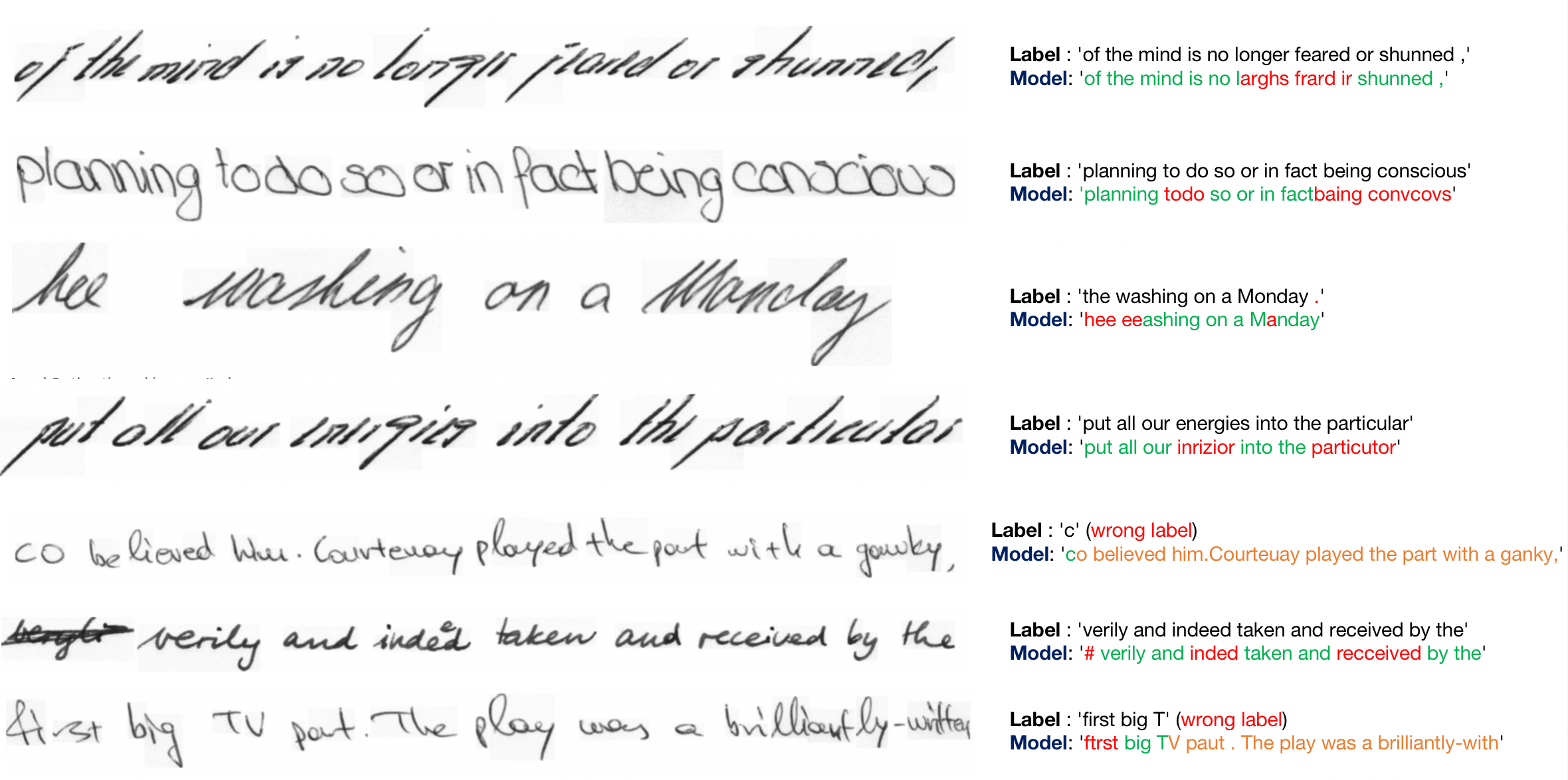}
  \caption{Examples where model makes mistakes}
  \label{fig:hard}
\end{figure*}

Figure \ref{fig:hard} shows some difficult examples (From IAM-Offline Test partition) where model makes mistakes. Sometimes these mistakes are due to noisy input, and sometimes even label is incorrect. We didn't explicitly correct any labels for our experiments. 

\end{document}